## ARTICLE INFORMATION

**Article title**

A Structured Dataset of Disease-Symptom Associations to Improve Diagnostic Accuracy

**Authors**

Abdullah Al Shafi[1], Rowzatul Zannat[1], Abdul Muntakim[1,*], Mahmudul Hasan[1]

**Affiliations**

[1] Khulna University of Engineering & Technology, Khulna-9203, Bangladesh

**Abstract**

Disease-symptom datasets are significant and in demand for medical research, disease diagnosis, clinical decision-making, and AI-driven health management applications. These datasets help identify symptom patterns associated with specific diseases, thus improving diagnostic accuracy and enabling early detection. The dataset presented in this study systematically compiles disease-symptom relationships from various online sources, medical literature, and publicly available health databases. The data was gathered through analyzing peer-reviewed medical articles, clinical case studies, and disease-symptom association reports. Only the verified medical sources were included in the dataset, while those from non-peer-reviewed and anecdotal sources were excluded. The dataset is structured in a tabular format, where the first column represents diseases, and the remaining columns represent symptoms. Each symptom cell contains a binary value, indicating whether a symptom is associated with a disease. Thereby, this structured representation makes the dataset very useful for a wide range of applications, including machine learning-based disease prediction, clinical decision support systems, and epidemiological studies. Although there are some advancements in the field of disease-symptom datasets, there is a significant gap in structured datasets for the Bangla language. This dataset aims to bridge that gap by facilitating the development of multilingual medical informatics tools and improving disease prediction models for underrepresented linguistic communities. Further developments should include region-specific diseases and further fine-tuning of symptom associations for better diagnostic performance.

## SPECIFICATIONS TABLE

| **Subject** | Computer Science |
| --- | --- |

| Specific subject area | Medical Informatics, Biomedical Data Science, Public Health |
|---|---|
| Type of data | Table(Processed) |
| Data collection | Very deep research regarding various online articles, medical literature, health-related blogs, and forums like [1] [2], including publicly available databases on health care, was employed in collecting the data. Only articles from internationally recognized medical journal outlets, well-documented cases of clinical observations, and all reports on disease-symptom cases were analyzed and extracted for pertinent information. Inclusion criteria focused on verified medical sources, while non-peer-reviewed or anecdotal reports were excluded. Then, the retrieved data were tabulated into a uniform table format and presented for comprehensiveness. |
| Data source location | Publicly available online sources, medical literature, and health databases. |
| Data accessibility | Repository name: Mendeley Data<br>Data identification number: 10.17632/rjgjh8hgrt.5<br>Direct URL to data: https://data.mendeley.com/datasets/rjgjh8hgrt/5 |
| Related research article | none |

## VALUE OF THE DATA

- The dataset provides a structured representation of relationships between diseases and symptoms. This contains the binary indicators (1-present, 0-absent) for every symptom analyzed in relation to a wide variety of diseases and thus will enable researchers to find patterns of co-occurrence and clusters of symptoms across many diseases. It helps to improve diagnostic accuracy and refines the symptom-based models for disease classification.
- The dataset will definitely help in training the models related to disease prediction, symptom analysis, and also clinical decision support systems using machine learning. Various algorithms can be developed based on this data for automated diagnosis, symptom-based disease prediction, and personalized medicine.
- Public health researchers can use this dataset to assess the prevalence of symptoms in different diseases, contributing to the early identification of new health trends and the formulation of strategies for health prevention. These data will further help in the syndromic surveillance system for early detection of outbreaks.

- This will serve as a good benchmark dataset on which newly developed methodologies will be tested in medical data analysis. The results can be cross-checked against this dataset for robustness, thus ensuring symptom-based disease identification with accuracy.
- This dataset can be very useful in research at several levels: computational biology, biomedical informatics, and artificial intelligence. The uniform structure provides better interoperability with other datasets from healthcare; therefore, enabling research across disciplines is possible.

## BACKGROUND

Systematic mapping between diseases and symptoms is critical in both basic research and clinical practice. These accurate datasets help enhance diagnostic accuracy, support clinical decision systems, and enable personalized treatment. Zlabinger et al. [3] proposed the Disease-Symptom Relation (DSR) collection, which provides symptom judgments in grades for diseases. Then, Electronic Health Records (EHRs) have further improved disease-symptom correlations [4], offering large-scale patient data for analysis. Sneha Grampurohit et al. [5] and Md. Atikur et al. [6] used a Kaggle dataset [7] of 4,920 records for predicting 41 diseases. Furthermore, M. M. Rahman et al. [8] translated this dataset into Bangla using the Google Translation API and thus made it usable for research in localized disease prediction. Even then, Bangla medical datasets are few and far between. Most of the medical records are unstructured, and the non-existence of standardized terminology further aggravates the problem of dataset creation. The limited digitization of health data in Bangladesh also makes the extraction of quality disease-symptom relationships a difficult task. This work, therefore, tries to fill this gap by compiling a structured dataset that collates information from varied medical sources. The dataset can support machine learning applications, epidemiological studies, and AI-driven diagnostics. This work will address the need for structured medical data in Bangla, hence promoting digitization, encouraging standardized medical terminology, and improving healthcare accessibility, disease prediction, and localized medical informatics.

## DATA DESCRIPTION

The dataset is in the form of a table and it shows a relationship between disease and symptom. It is organized so that the leftmost column is diseases, and the rest are symptoms. Every cell contains a binary value (1 or 0), where:

- 1 signifies that the symptom shares a relation with the disease.
- 0 indicates no relationship.

The following files are available in our data repository [9]:

- **dataset.xlsx**: This dataset contains 1 if a symptom is related to the disease and the remaining cells are kept blank.
- **cleaned_dataset.xlsx**: The complete dataset after the cleaning processes described in the next section.
- **cleaned_dataset_with_english_translation.xlsx**: The cleaned dataset with the English translation of all the diseases and symptoms.

**Table 1**

Overview of our dataset

| Attribute | Total |
|---|---|
| Unique Diseases | 85 |
| Unique Symptoms | 172 |
| Disease-Symptoms Relation | 758 |

Table 1 presents the quantitative overview of our suggested dataset, emphasizing the number of unique diseases, unique symptoms, and the total number of disease-symptom relationships (data volume). The total disease-symptom relationships are the sample size, showing the vast amount of relationships included in the dataset. This information is crucial to comprehend the comprehensiveness of the dataset and its potential for effective predictive modeling.

| Diseases | | |
|---|---|---|
| ফাঙ্গাস (Fungal infection) | ডায়েবেটিস (Diabetes) | ক্যান্সার (Cancer) |
| ডিমনেশিয়া (Dementia) | স্ট্রোক (Stroke) | ডেঙ্গু (Dengue) |
| আমাশয় (Dysentery) | জলবসন্ত (Chickenpox) | কোষ্ঠকাঠিন্য (Constipation) |
| চোখ ওঠা (Conjunctivitis) | গুটিবসন্ত (Smallpox) | দাদ রোগ (Ringworm) |
| বাত (Gout) | কলেরা (Cholera) | নিমোনিয়া (Pneumonia) |
| ম্যালেরিয়া (Malaria) | হার্ট অ্যাটাক (Heart Attack) | বধিরতা (Deafness) |
| পারকিনসন্স (Parkinson's Disease) | গ্লুকোমা (Glaucoma) | ছানি (Cataract) |
| বেরিবেরি (Beriberi) | রিকেট (Rickets) | জলাতঙ্ক (Rabies) |
| সোয়াইন ফ্লু (Swine Flu) | পক্ষাঘাত (Paralysis) | এপিলেপসি (Epilepsy) |
| নিপাহ ভাইরাস (Nipah Virus) | ধনুষ্টঙ্কার (Tetanus) | ওটিটিস মিডিয়া (Otitis Media) |
| গ্লসাইটিস (Glossitis) | কুষ্ঠ (Leprosy) | চিকুনগুনিয়া (Chikungunya) |
| আলসার (Ulcer) | ইনফ্লুয়েঞ্জা (Influenza) | হাম (Measles) |
| হিমোফিলিয়া (Hemophilia) | স্কার্ভি (Scurvy) | প্লুরিসি (Pleurisy) |
| পায়োরিয়া (Pyorrhea) | ট্র্যাকোমা (Trachoma) | সাইনোসাইটিস (Sinusitis) |
| ব্রংকাইটিস (Bronchitis) | পোলিও (Polio) | স্ক্লেরোসিস (Sclerosis) |
| মস্তিষ্কের টিউমার (Brain Tumor) | এনসেফালাইটিস (Encephalitis) | সিসিএইচএফভি (CHIKV) |
| লিউকোমিয়া (Leukemia) | ডিপথেরিয়া (Diphtheria) | যক্ষ্মা (Tuberculosis) |
| টাইফয়েড (Typhoid) | জন্ডিস (Jaundice) | গলগণ্ড (Goiter) |
| রিফট ভ্যালি ফিভার (Rift Valley Fever) | এলার্জি (Allergy) | জিইআরডি (GERD) |
| ক্রনিক কোলেস্টেসিস (Chronic Cholecystitis) | ওষুধের প্রতিক্রিয়া (Drug Reaction) | পেপটিক আলসার (Peptic Ulcer) |
| এইডস (AIDS) | গ্যাস্ট্রোএন্টেরাইটিস (Gastroenteritis) | ব্রঙ্কিয়াল অ্যাজমা (Bronchial Asthma) |
| উচ্চ রক্তচাপ (Hypertension) | মাইগ্রেন (Migraine) | সার্ভিকাল স্পন্ডাইলোসিস (Cervical Spondylosis) |
| প্যারালাইসিস (Paralysis) | হেপাটাইটিস এ (Hepatitis A) | হেপাটাইটিস বি (Hepatitis B) |
| হেপাটাইটিস সি (Hepatitis C) | হেপাটাইটিস ডি (Hepatitis D) | হেপাটাইটিস ই (Hepatitis E) |
| অ্যালকোহলিক হেপাটাইটিস (Alcoholic Hepatitis) | সাধারণ সর্দি (Common Cold) | পাইলস(Piles) |
| ভ্যারিকোস ভেইন (Varicose Veins) | হাইপোথাইরয়েডিজম (Hypothyroidism) | হাইপারথাইরয়েডিজম (Hyperthyroidism) |
| হাইপোগ্লাইসেমিয়া (Hypoglycemia) | অস্টিওআর্থারাইটিস (Osteoarthritis) | ভারটিগো (Vertigo) |
| ব্রণ (Acne) | মূত্রনালীর সংক্রমণ (Urinary Tract Infection) | সোরিয়াসিস (Psoriasis) |
| ইমপেটিগো (Impetigo) | | |

**Fig 1.** List of the diseases in proposed datasets.

Fig. 1 gives a full list of all 85 diseases in our dataset. It covers many different medical conditions. We included a wide range of diseases, like infectious diseases, chronic conditions, and rare disorders. This mix makes sure our dataset is representative of many types of illnesses. Having this variety helps machine learning models that use this dataset to perform well with different medical conditions.

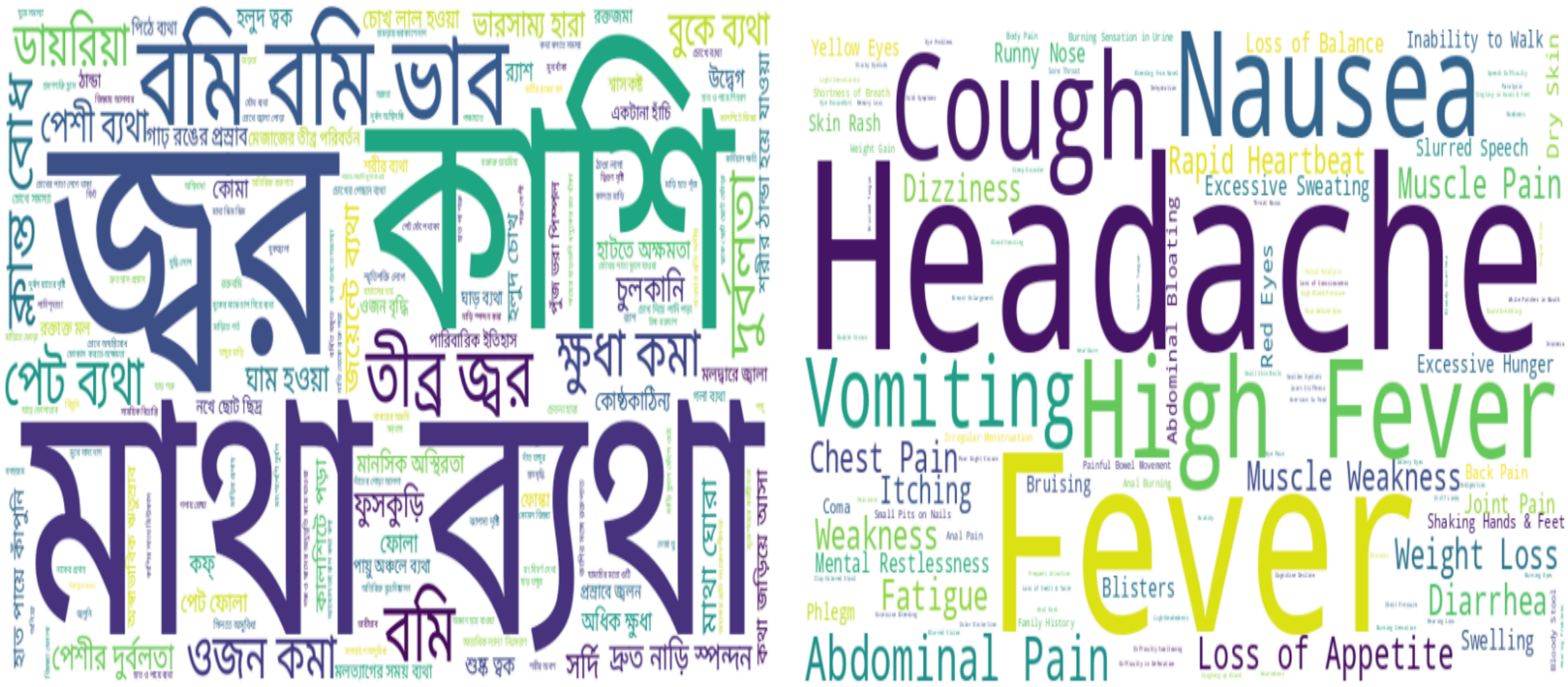

**Fig 3.** Word Cloud of symptoms in our dataset (left: in Bengali and right: in English).

The word cloud in Fig. 3 is an illustrative display of symptom frequency and distribution in our dataset. Symptoms are listed on the left in Bengali and on the right in English. More frequent symptoms are larger and more intensely formatted in the word cloud, so you can quickly see which symptoms recur most frequently. There are 172 various symptoms, displaying numerous potential signs of various diseases in our data. The size of words varies with symptom frequency—larger words indicate they occur more frequently. This tool assists us in comprehending the diversity of symptoms, displaying that numerous conditions are taken into account. As there is a wide range of symptoms, the dataset is witnessing many diseases, indicating that all these symptoms need to be studied to predict and diagnose diseases.

Some symptoms are very common in many diseases, and they can be termed common symptoms. Fig. 4 shows the top 20 most common symptoms. The most common is মাথাব্যথা (Headache) with 156 cases. Others include বমি বমি ভাব (Nausea) with 145 cases, বমি (Vomiting) with 144 cases, ক্লান্ত বোধ (Fatigue) with 140 cases, and তীব্র জ্বর (High Fever) with 131 cases. These symptoms occur in various diseases and are thus highly significant for diagnosis but are not specific. Respiratory signs like কাশি (Cough) with 121 cases and শ্বাসকষ্ট (Shortness of Breath) with 117 cases can reflect issues like pneumonia, asthma, or viral infections. Gastrointestinal signs, including পেট ব্যথা (Stomach Pain), ডায়রিয়া (Diarrhea), and ক্ষুধা কমা (Loss of Appetite), are also common. Because many diseases share these symptoms, prediction models must take into account how symptoms occur together, and not individually. To improve accuracy, less common but specific symptoms for certain diseases should also be included. The high occurrence of general symptoms makes it challenging to classify diseases, requiring advanced techniques in predictive modeling.

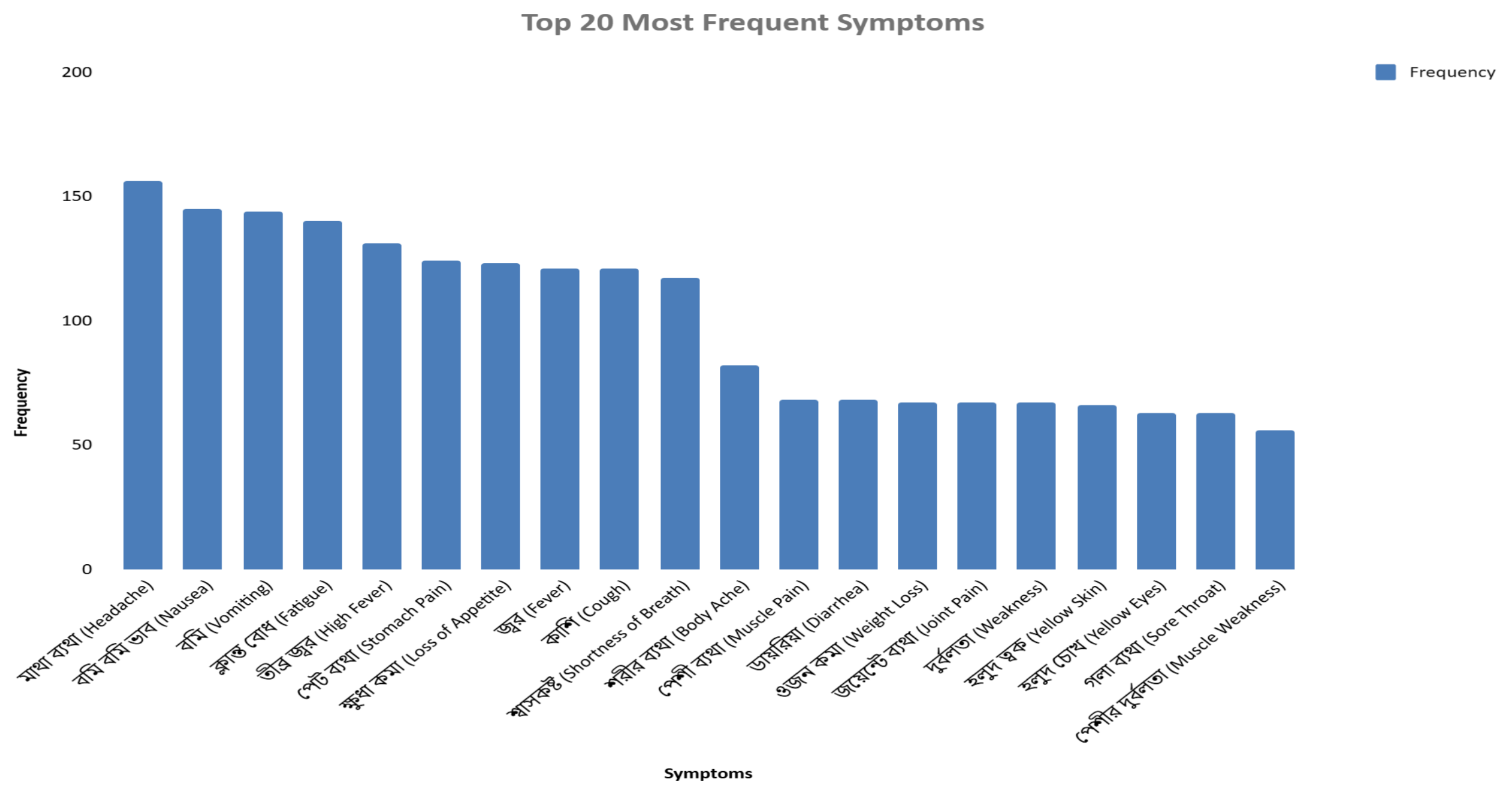

**Fig 4.** Most frequent symptoms.

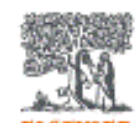

Fig. 5 shows a sample portion of our proposed dataset. Each row represents a separate disease, and the columns indicate the symptoms seen in patients with these diseases. There are 172 symptoms listed in all, each of them measured as either a 0 or a 1. A 1 would mean the symptom has been seen, and a 0 would mean that it hasn't been seen. For instance, the symptom শ্বাসকষ্ট (Shortness of breath) is assigned a 1 if present in a particular disease and a 0 otherwise. As numerical as the data are presented, the majority are categorical, diagnosing diseases based on qualitative descriptions of symptoms.

| Symptoms | | | | | | | | | | | | | Disease |
|---|---|---|---|---|---|---|---|---|---|---|---|---|---|
| শ্বাসকষ্ট | তীব্র জ্বর | পেট ব্যথা | ... | বমি | দুর্বলতা | শরীর ব্যথা | ফুসকুড়ি | র‍্যাশ | অজ্ঞান | বুকে ব্যথা | কাশি | অজ্ঞান | |
| 1 | 1 | 0 | ... | 0 | 0 | 0 | 0 | 1 | 0 | 0 | 0 | 1 | ডেঙ্গু (Dengue) |
| 0 | 0 | 1 | ... | 1 | 1 | 0 | 0 | 0 | 0 | 0 | 0 | 0 | আমাশয় (Dysentery) |
| 1 | 1 | 0 | ... | 0 | 0 | 1 | 0 | 0 | 0 | 1 | 1 | 0 | নিমোনিয়া (Pneumonia) |
| 0 | 1 | 0 | ... | 0 | 1 | 1 | 0 | 0 | 0 | 0 | 0 | 0 | ম্যালেরিয়া (Malaria) |
| 1 | 1 | 0 | ... | 0 | 0 | 1 | 0 | 0 | 0 | 1 | 1 | 0 | ব্রংকাইটিস (Bronchitis) |

**Fig 5.** Snapshot of the dataset

## EXPERIMENTAL DESIGN, MATERIALS AND METHODS

The dataset consists of 172 binary features representing the presence or absence of symptoms for various diseases. Fig. 6 shows the whole procedure to develop the dataset. The steps to develop the dataset are given below:

- **Data collection and annotation**

  The raw dataset is stored in the spreadsheet file "dataset.xlsx" and contains symptom-based diagnostic data for various diseases. The dataset was created by systematically mapping symptoms to diseases based on clinical observations from online resources like- Bangla blogs, Bangla newspapers, online surveys, expert medical knowledge, and available diagnostic guidelines. The construction process involved the following key steps:

  1. **Symptom Selection:**
     - A list of common and clinically relevant symptoms was curated from medical literature and expert consultations.
     - Each symptom was chosen based on its association with multiple diseases, ensuring broad coverage.
  2. **Disease Mapping:**
     - A set of diseases was selected, each associated with a distinct symptom pattern.
     - Symptoms were manually assigned to diseases based on medical diagnosis criteria.
  3. **Binary Encoding:**
     - Each symptom was assigned the value 1 if it appeared commonly in a disease and 0 otherwise.
     - The information was tabulated into a binary matrix where rows represent diseases and columns represent symptoms.

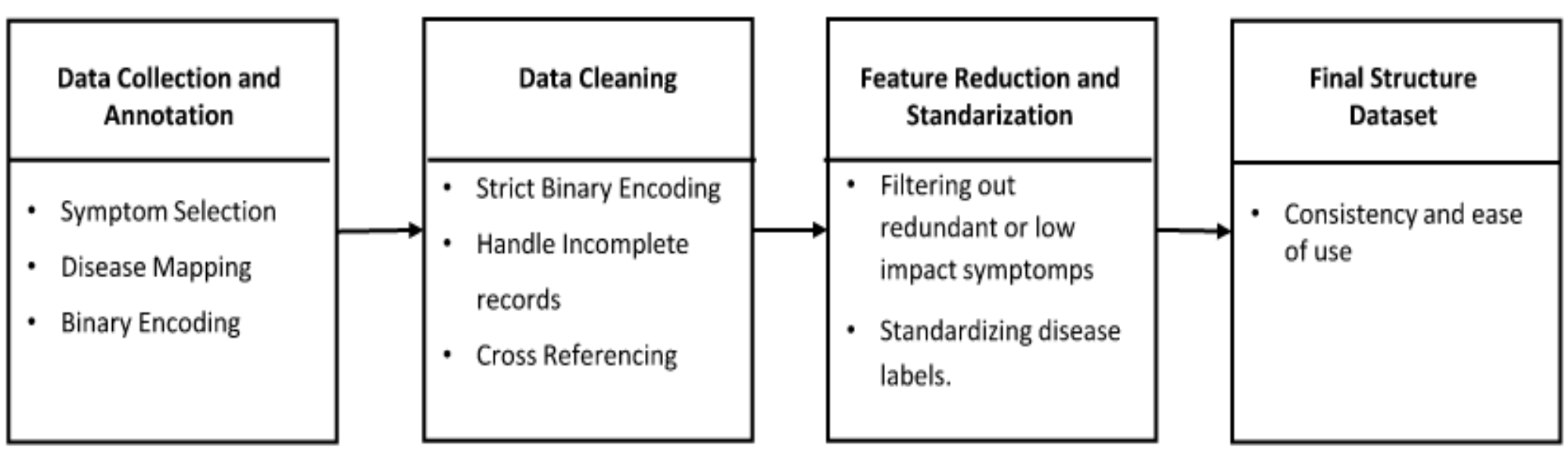

**Fig 5.** The procedure to develop the proposed dataset.

- **Data Cleaning**

The foremost thing that was done was to ensure that the encoding for all symptoms was strictly binary. The symptoms were originally recorded in a variety of formats, so data standardization was important so that 1 indicated the presence of a symptom while 0 indicated its absence from a symptom. This required careful examination of the raw data for detection and correction of any inconsistent or non-binary values. It was found that some records contained values other than 0 or 1, which were corrected to conform to the binary structure or removed entirely to maintain the integrity of the dataset.

Apart from maintaining binary consistency, incomplete records were addressed by filling in gaps where practicable or deleting the records in their entirety if missing data could not be estimated reasonably. This helped maintain a clean dataset of unambiguous or reliable information.

Moreover, various symptom mappings were found to be inconsistent. For example, some diseases included symptoms that had no reference variable per the medical guidelines. This was corrected by a detailed manual review whereby each disease's symptoms were cross-checked from genuine medical resources to ensure validity. Since symptoms of the diseases inconsistently mapped against each other were thus updated, valid and consistent symptom-disease relationships will be reflected in the data.

- **Feature Reduction and Standardization**

Once the cleaning was completed, the next was to aim at eliminating the redundant or low-impact symptoms. Some symptoms appeared in almost all diseases, hence being irrelevant in distinguishing between them. Similarly, some symptoms appeared in very few diseases to be of any use in prediction. These were either eliminated or combined, so only relevant symptoms remained in the dataset. This selection procedure reduced noise and improved the overall accuracy of the classification models.

Standardization of the labels for the diseases constituted an important step in the cleansing. It involved the elimination of any disparity in the names of the diseases from the names. This involved eliminating the synonyms, spelling differences, as well as format differences. By giving the same uniform and consistent label to all the diseases, we nullified any disparity in writing the diseases into the dataset. Standardization allows easier work with data and prevents interpretation or training of errors in models.

To evaluate how effective the proposed dataset is, varied machine learning models were tested. They varied from Perceptron, Logistic Regression, and Naive Bayes to Decision Trees, K-Nearest Neighbors, Passive Aggressive Classifiers, Random Forest, and Support Vector Machines. The comparison of models for disease classification was thus done on the basis of accuracy, precision, recall, and F1-measure. The accuracy was found to be highest (0.97) for Logistic Regression, Perceptron, and Random Forest. The best model precision, recall, and F1-score were provided by Logistic Regression, followed by Random Forest. In contrast, Decision Tree came out to have poor performance in all aspects, with its accuracy being 0.78 and low F1-scores. Some models, such as K-Nearest Neighbors and Passive Aggressive Classifier, had a good performance on disease classification and comparative stability throughout. This supports that the dataset is amenable to classification tasking and has stability for diverse machine learning applications.

**Table 3**

Performance of various Machine Learning models on disease classification with the proposed dataset [10]:

| Model | Accuracy | Precision | Recall | F1-Score |
|---|---|---|---|---|
| Perceptron | 0.97 | 0.93 | 0.92 | 0.92 |
| Logistic Regression | 0.97 | 0.96 | 0.95 | 0.95 |
| Naive Bayes | 0.94 | 0.91 | 0.90 | 0.90 |
| Decision Tree | 0.78 | 0.71 | 0.70 | 0.69 |
| K-Nearest Neighbors | 0.96 | 0.93 | 0.91 | 0.91 |
| Passive Aggressive Class | 0.96 | 0.91 | 0.90 | 0.90 |
| Random Forest | 0.97 | 0.96 | 0.96 | 0.96 |
| Support Vector Machine | 0.96 | 0.94 | 0.92 | 0.92 |

## LIMITATIONS

Though an incredibly wide range of diseases and symptoms is included within the dataset, it may well not be complete. There may be very few very uncommon disorders and local diseases that are underrepresented, as little published literature is present. The information is not in real-time, i.e., it does not update automatically from time to time to accommodate recent medical developments or new illnesses.

## ETHICS STATEMENT

No direct interaction with patients or healthcare providers was conducted during data collection. As the dataset is derived from secondary sources, ethical approval, and informed consent were not required. All sources used to comply with open-access policies or are publicly available for research purposes. Additionally, efforts were made to ensure that the dataset does not contain any sensitive or confidential patient data.

## CRediT AUTHOR STATEMENT

**Abdullah Al Shafi**: Conceptualization, Data curation, Visualization, Validation, Writing –Original draft; **Rowzatul Zannat**: Methodology, Data curation, Investigation, Writing – original draft; **Abdul Muntakim**: Supervision, Validation, Writing –review and editing; **Mahmudul Hasan**: Writing –review and editing

## ACKNOWLEDGEMENTS

We would like to express our gratitude to the researchers and medical professionals whose publicly available datasets, studies, and articles contributed to the development of this dataset. Special thanks to the contributors of open-access medical literature and health databases that provided valuable insights into disease-symptom relationships. This research did not receive any specific grant from funding agencies in the public, commercial, or not-for-profit sectors.

## DECLARATION OF COMPETING INTERESTS

The authors declare that they have no known competing financial interests or personal relationships that could have appeared to influence the work reported in this paper.